\documentclass{ifcolog1}
\title{Relating Information and Proof}
\titlerunning{Relating Information and Proofs }

\addauthor[slissenko@u-pec.fr]
{Anatol Slissenko}
{Laboratory for Algorithmics, Complexity and Logic (LACL)\protect \\
 University Paris East Cr\'eteil (UPEC) \\
 61 av. du G\'en\'eral de Gaulle 94010, Cr\'eteil France}
\authorrunning{Anatol Slissenko}


\titlethanks{}

  \def\Pr{\bmth{P}}    
  \def\ES{\emptyset}          
  \def\QBu{\bmth{QB1}}
  \def\QBd{\bmth{QB2}}
  \def\QBt{\bmth{QB3}}
  \def\QDu{\bmth{QD1}}
  \def\QDd{\bmth{QD2}}
  \def\QDt{\bmth{QD3}}
  \def\QFu{\bmth{QF1}}
  \def\QB{\bmth{QB}}
  \def\QD{\bmth{QD}}
 \newcommand{\bmth}[1]{\mbox{\boldmath $#1$}}
 \newcommand{\Hc}[1]{\hspace*{#1cm}}
 \newcommand{\NBF}[1]{\noindent{\bf #1}}
  
 \newtheorem{prop}{Proposition}

 \def\IM{\rightarrow}

 \def\NI{\noindent}
 
 \def\CA{{\mathcal{A}}}
 
 \def\CD{{\mathcal{D}}}
 \def\CF{{\mathcal{F}}}
 \def\CG{{\mathcal{G}}}
 \def\CP{{\mathcal{P}}}
 \def\CQ{{\mathcal{Q}}}
 \def\CS{{\mathcal{S}}}

 \def\alI{\mathit{\alpha}}
 \def\beI{\mathit{\beta}}
 
 \def\deI{\mathit{\delta}}
 \def\zeI{\mathit{\zeta}}
 \def\phI{\mathit{\phi}}
 \def\PhI{\mathit{\Phi}}

\usepackage{latexsym}
\usepackage{amsmath}
\usepackage{amssymb}

\begin{document}
\maketitle

\begin{abstract}
In mathematics information is a number  that measures uncertainty (entropy) based on a probabilistic distribution, often of an obscure origin. In real life language information is a datum, a statement, more precisely, a formula.
But such a formula should be justified by a proof. I try to formalize this perception of information. The measure 
of informativeness of a proof is based on the set of proofs related to the formulas under consideration.  This set of possible proofs (`a knowledge base') defines a probabilistic measure, and entropic weight is defined using this measure. The paper is mainly conceptual, it is not clear where and how this approach can be applied.
\end{abstract}

 \section{Introduction}  
  One can see that the meanings of the word "information" in real life (in colloquial speech) and in mathematics have  little in common, to put it mildly. Information in colloquial speech is a datum, or more precisely a statement like "$X$ is a winner of a competition $C$", and in mathematics information is a number based on an evaluation of chances, often very personal. It fact, in mathematics it is entropy, a measure of uncertainty based on a probabilistic measure.
   
   This discrepancy is well known and was a subject of discussions by philosophers, e.g., 
   \cite{Floridi2013,Adriaans2013}.

   In this paper I describe a piece of information as a formula, and its quantity is evaluated on the basis 
 of the knowledge  that permits to prouve the formula. A probabilistic distribution is used but it is defined
 by the knowledge, and as a measure I take what is called below entropic weight that has a flavor of entropy. 
     
 \section{Motivational example}
 
  Three men named \emph{Bok}, \emph{Dok}, \emph{Fok} participate in a competition \textbf{C} where there is only one winner. The winner is announced by different sources at the same time (with this assumption we avoid mentioning the time moments).
  
  Look at the classical view at the quantity of information received by different persons. 
  
  Someone called $X$ estimates that\\
  \emph{Bok} wins with probability $\frac{1}{4}$,\\  
  \emph{Dok} wins with probability $\frac{1}{4}$,\\
  \emph{Fok} wins with probability $\frac{1}{2}$.\\
  
   Someone called $Y$ estimates that\\   
  \emph{Bok} wins with probability $\frac{1}{8}$,\\  
  \emph{Dok} wins with probability $\frac{7}{16}$,\\
  \emph{Fok} wins with probability $\frac{7}{16}$.\\
  
  Someone called $Z$ has no estimations of chances at all, so for him all outcomes are equiprobable. \\
  
  In mathematics the information of $X$ about the winner is 
  
  \NI$-(\frac{1}{4}\log\frac{1}{4}+\frac{1}{4}\log\frac{1}{4}+\frac{1}{2}\log\frac{1}{2})=\frac{3}{2}=1.5$,
  
  \NI{} the information of $Y$  about the  winner is 
  
  \NI$-(\frac{1}{8}\log\frac{1}{8}+\frac{7}{16}\log\frac{7}{16}+\frac{7}{16}\log\frac{7}{16})\approx1.42$ 
 
  \NI{} the information of $Z$  about the  winner is 
  
  \NI$-(3\cdot\frac{1}{3}\log\frac{1}{3})=\log3\approx1.58$

  Do the numbers $1.5$, $1.4$ and $1.58$ give information about the winner we are interested in? 
  Surely, not.  They evaluate the uncertainty of the systems involved,  and these uncertainties are individual. 
  
  Suppose that \emph{Bok} is the winner. Suppose that \emph{in some way} this information was received by 
  $X$, $Y$ and $Z$. It is the same for all of them. 
  
   What is "Bok is the winner" mathematically?  It is a logical formula, namely, "Bok is the winner of competition $\bmth{C}$". 
  
  How do we get this information (that is, clearly, not a number)? 
  
  "Bok is the winner" is of value if it comes with a \emph{proof} that \emph{Bok} is the winner. Such a proof may be of the following kind.
  
  "Radio station $\bmth{R}$ always gives truthful information about competitions.\\ 
  It broadcasted that \emph{Bok} is the winner of the competition". 
  
  This is the information we are interested in. Look at it more formally.\\ 

  \subsection{Inference system and proofs for the example}\label{subsInfSysExmp}   
  \NBF{Constants.}
  
   $\CP=\{Bok,Dok,Fok\}$ are participants, $\CS=\{R_1,R_2,R_3\}$ is a finite set of information (data) sources (some of them are always truthful, others are always deceitful or may be sometimes truthful, sometimes  deceitful.)
  
  \NBF{Predicates and functions.}
    
  $Day$ is the name of the day we speak about. 

  $Win(\alI)$, where $\alI\in\CP$, says that $\alI$ is the winner
  
  $Brd(R,\alI)$ says that $R$ broadcasts that $\alI$ is the winner. 
  
  $Fri$ is an abbreviation of $Friday$, the name of a particular day of the week.
       
  \NBF{Axioms.}
  
  If $R\in\CS$ is truthful and $R$ broadcasts $\PhI$  then $\PhI$.
  
  If $R$ is deceitful and $R$ broadcasted $\PhI$ then $\neg\PhI$.
  
  Source $R_1$ is always truthful. 
  
  Source $R_2$ is truthful on Fridays and deceitful on other days.  
  
  Source $R_3$ is always deceitful.
  
  There is a always winner : $(Win(Bok)\vee{}Win(Dok)\vee{}Win(Fok))$.
  
  There is at most one winner : $(Win(\alI)\IM\neg{}Win(\beI)$ for $\alI\neq\beI$.
  
  \NBF{User's Data (possible user's axioms).}    
  
  $Brd(R,\alI)$, i.e., $R$ broadcasts that $\alI$ is a winner, where $R\in\CS$ and $\alI\in\CP$.
  
  $Day(t)=Fri$. 
  
  \vspace{1ex}  
    As inference rules we use  axioms and predicate logic .   
  
  \NBF{Proofs.}
  
  Any proof starts with user's data. 
  
  If a user says something about broadcast, then for the user it is truthful. We assume that
  in the set of proofs (that is a `knowledge system') all the proofs have the final formula of 
  the  form  $Win(\alI)$, though the inference system outlined above permits proofs with other final formulas,
  in particular  like $Win(Bok)\vee{}Win(Fok)$.
  
  \vspace{1ex}  
  Here is a set of proofs that that are presumed to constitute a `knowledge system'. 
  We do not make explicit the analysis, i.e., how this or that formula is obtained, it is evident.
  
  \vspace{1ex}   
  
 \NI$\QBu$: $Day=Fri$, $Brd(R_2,Bok)$, $Win(Bok)$.
 
 \NI$\QBd$: $Day\neq{}Fri$, $Brd(R_2,Dok)$, $Brd(R_1,Bok)$, $Win(Bok)$.
  
 \NI$\QBt$: $Day\neq{}Fri$, $Brd(R_2,Dok)$, $Win(Bok)\vee{}Win(Fok)$, $Brd(R_3,Fok)$, 
 
 \Hc{0.5}$\neg{}Win(Fok)$, $Win(Bok)$.  
  
 \NI$\QDu$:  $Brd(R_1,Dok)$, $Win(Dok)$.
    
 \NI$\QDd$: $Day=Fri$, $Brd(R_3,Fok)$,   $Brd(R_2,Dok)$, $Win(Dok)$. 
        
 \NI$\QDt$: $Day=Fri$, $Brd(R_3,Fok)$, $Brd(R_1,Dok)$,  $Brd(R_2,Dok)$, $Win(Dok)$. 
 
 \NI$\QFu$: $Day\neq{}Fri$, $Brd(R_2,Dok)$, $Win(Bok)\vee{}Win(Fok)$, $Brd(R_3,Bok)$,
  
 \Hc{0.5}$\neg{}Win(Bok)$, $Win(Fok)$.  
        
  \section{Informativeness of Proofs}
  
  How much of information one have in a proof? First, we describe a  possible approach
  in terms related to the example (and to logic), and after that in section~\ref{EntWghtAbstr} 
  we give an abstract set-theoretic framework that do not mention logic. 
  
  The proofs we consider are proofs of responses to information queries. Any information query is something like
  $\bmth{Find}\,x\,\PhI(x)$ where $\PhI(x)$ is a formula, an information property. 
  An answer to such a query is a formula
  $\PhI(\alI)$ with a constant $\alI$. And an information proof is proof of $\PhI(\alI)$. 
  
  A knowledge system is a set of information proofs. In our setting all sets are finite.  
  
  As compared to probabilistic distributions, often with obscure origin, used in the evaluation of entropy, 
  the knowledge system in our approach can be shared by all individuals involved. 
  And the probabilistic measure that we use
   depends only on this knowledge system and not on individual vision of the situation.
  
  \subsection{Entropic Weight}\label{EntWght}
  
  Notations: 
  
  $\bullet$ $\CA$ is a set of constants that are used in answers to queries. 
  
  $\bullet$ $\CF=\{\PhI(\alI)\}_{\alI}$ is a set of answers.
   
  $\bullet$ $\CQ$ is the set of all proofs.
  
  $\bullet$ $\CQ_{\PhI(\alI)}=\CQ_{\alI}=\{Q\in\CQ: Q \textrm{\;is a proof\; of\;}\PhI(\alI)\}$  
      $\CQ_{\alI}$ is a set of proofs of $\PhI(\alI)$.
      
  $\bullet$ $q_{\alI}=|\CQ_{\alI}|$, $M=|\CA|$.
  
  For simplicity we assume that for a given $\alI$ there is exactly one $\PhI(\alI)$  but it may have many
  different proofs.
  
  $\bullet$  Probabilistic measure $\Pr$ on the proofs:  $\Pr(\CQ_{\alI})=\frac{1}{M}$,
  $\Pr(Q)=\frac{1}{M\cdot{}q_{\alI}}$ for $Q\in\CQ_{\alI}$.
  
  The measure is based on the \emph{principle of maximum uncertainty} that says that all answers are 
  equiprobable, and for a given answer all its proofs are also equiprobable. Under this assumption the
  uncertainty is maximal.
  
  A proof consists of formulas and of analysis, i.e., of references to the rules applied, 
  but  below we treat it as just  a set  of formulas. 
  
  We wish to measure `informativeness' of a given proof. In other words, how one gets more and more information
  by obtaining bigger and bigger subsets of the proof.
   
  To do it, for a given subset of a given proof  we introduce entropic weight -- 
  a measure with a flavor of entropy that has properties corresponding
  to the intuition in the context under consideration.
  
  For a subset $S$ of formulas of a proof we set $E(S)=\{Q:S\subseteq{}Q\}$, and define its
   \emph{entropic weight} $\CD(S)$:
   
   \begin{equation}\label{CD1}
\CD(S)=\CD(E(S)=
  -\sum_{\alI}\Pr(E(S)\cap\CQ_{\alI})\log\frac{\Pr(E(S)\cap\CQ_{\alI})}{\Pr(E(S))},
    \end{equation}
 here and below $\log$ is $\log_2$.
 
  Taking into account that the sets $(E(S)\cap\CQ_{\alI}$ are disjoint, $\bigcup_{\alI}\CQ_{\alI}=\CQ$, 
  $\Pr(\CQ)=1$, and thus $\sum_{\alI}\Pr(E(S)\cap\CQ_{\alI})=\Pr(E(S))$ 
  we can rewrite formula (\ref{CD1}) for $\CD(S)$ as:\\
    \begin{equation}\label{CD2}
\CD(S)=-\sum_{\alI}\Pr(E(S)\cap\CQ_{\alI})\log\Pr(E(S)\cap\CQ_{\alI})+\Pr(E(S))\log\Pr(E(S))
    \end{equation}
  Notice that the notation $\CD(S)$  with argument $S$, and not $E(S)$, is in a way misleading: 
  when $S$ grows the argument $E(S)$, 
  that is in fact used, grows down (non strictly). 
  
  Entropic weight $\CD(S)$ has the following properties:
  
  (D1) $\CD(\ES)=\log{}M$ (maximal uncertainty)
  
  (D2) $\CD(S)=0$ for any $\alI\in\CA$ and any $S\subseteq\CQ_{\alI}$ such that $E(S)\subseteq\CQ_{\alI}$
   
  \Hc{0.9}(maximal certainty)
  
  (D3) $\CD(S)$ is non-increasing when $S$ grows: 
      if $S\subseteq{}S'$ then $\CD(S)\geq\CD(S')$ 
      
 \Hc{0.9}(the uncertainty does not grow with getting more and more of information). 

\vspace{1ex}            
For the proof  see Proposition~\ref{CDprop} in section~\ref{EntWghtAbstr} below. 
      
  In order to evaluate evolution of informativeness  we look at what happens with entropic weight when 
  the size of subsets $S$ grows. How to choose subsets? We do it again according the principle of maximal uncertainty. Imagine that the choice is being done by an adversary who tries to maximize the uncertainty. 
  
  Look at the example.
            
\subsection{Entropic weight for the example}\label{EntWghtExmp} 
   
  \NI{}The measure $\Pr$ of each proof $\QB{}\bmth{i}$, $\QD{}\bmth{i}$, $1\leq{}i\leq3$, is $\frac{1}{9}$, 
   and that of $\QFu$ is $\frac{1}{3}$; 
   
   $\CQ_{Bok}=\{\QB\bmth{i}\}_{i=1,2,3}$, $\CQ_{Dok}=\{\QD{}\bmth{i}\}_{i=1,2,3}$, 
   $\CQ_{Fok}=\{\QFu\}$.\\  
   
 Consider proof $\QBu$. For one-element subset $U_0=\{Brd(R_2,Bok)\}$ or $U_0=\{Win(Bok)\}$  we have 
 $E(U_0)=\{\QBu\}$ and $\CD(U_0)=0$ as follows from (D2). Such a choice of one-element subset does not
 give maximal entropic weight for one-element subsets. 
 
 Take the remaining one-element subset, namely, $U_1=\{Day=Fri\}$. Then $E(U_1)=\{\QBu,\QDd,\QDt\}$, and
 $\Pr(E(U_1))=\frac{1}{9}+\frac{1}{9}+\frac{1}{9}=\frac{1}{3}$. 
 
 The measures of intersections are:
  $\Pr(E(U_1)\cap\CQ_{Bok})=\Pr(\{\QBu\})=\frac{1}{9}$, 
   
   $\Pr(E(U_1)\cap\CQ_{Dok})=\Pr(\{\QDd,\QDt\})=\frac{2}{9}$, 
   $\Pr(E(U_1)\cap\CQ_{Fok})=\Pr(\ES)=0$.
   
   With (\ref{CD2}) we get
   
   $\CD(U_1)=-\frac{1}{9}\log\frac{1}{9}-\frac{2}{9}\log\frac{2}{9}+\frac{1}{3}\log\frac{1}{3}\approx0.31$
 
   If we extend the set $U_1$ in any way to $U_1'$ we get $\CD(U_1')=0$ as follows from (D2). So for 
   1-element subsets of the proof $\QBu$ the maximal entropic weight is approximately $0.31$, and 
   for 2-element subsets 
   of the proof $\QBu$ the maximal entropic weight is $0$. The passage from $0.3$ to $0$ shows 
   the speed of convergence to complete certainty.
 
 Consider proof $\QBt$, and three sets 
 
   $S_1=\{Brd(R_2,Dok)\}$, 
 
   $S_2=\{Day\neq{}Fri,\,Brd(R_2,Dok)\}$,
   
   $S_3=\{Day\neq{}Fri,\,Brd(R_2,Dok),Win(Bok)\vee{}Win(Fok)\}$.
   
   These sets maximize the entropic weight for the sets of size respectively $1$, $2$, $3$.
   
 \NI{}We have  $E(S_1)=\{\QBd,\QBt,\QDd,\QDt,\QFu\}$,  
   $E(S_2)=\{\QBd,\QBt,\QFu\}$,\\
    $E(S_3)=\{\QBt,\QFu\}$, and $\Pr(E(S_1))=\frac{4}{9}+\frac{1}{3}=\frac{7}{9}$,    
   $\Pr(E(S_2))= \frac{2}{9}+\frac{1}{3}=\frac{5}{9}$, $\Pr(E(S_3))=\frac{1}{9}+\frac{1}{3}=\frac{4}{9}$.
   
 For intersections of $E(S_1)$ with $\CQ_{\alI}$ we have $E(S_1)\cap\CQ_{Bok}=\{\QBd,\QBt\}$,\\ 
 $E(S_1)\cap\CQ_{Dok}=\{\QDd,\QDt\}$, 
    $E(S_1)\cap\CQ_{Fok}=\{\QFu\}$, and 
    
    \NI$\Pr(E(S_1)\cap\CQ_{Bok})=\Pr(E(S_1)\cap\CQ_{Dok})=\frac{2}{9}$,
    $\Pr(E(S_1)\cap\CQ_{Fok})=\frac{1}{3}$.
    
 For intersections of $E(S_2)$ we have $E(S_2)\cap\CQ_{Bok}=\{\QBd,\QBt\}$, 
 
 \NI$E(S_2)\cap\CQ_{Dok}=\ES$,
   $E(S_2)\cap\CQ_{Fok}=\{\QFu\}$, and 
   
   \NI$\Pr(E(S_2)\cap\CQ_{Bok})=\frac{2}{9}$, $\Pr(E(S_2)\cap\CQ_{Dok})=0$,
    $\Pr(E(S_2)\cap\CQ_{Fok})=\frac{1}{3}$.
    
 For intersections of $E(S_3)$ we have $E(S_3)\cap\CQ_{Bok}=\{\QBt\}$, $E(S_3)\cap\CQ_{Dok}=\ES$,\\ 
    $E(S_3)\cap\CQ_{Fok}=\{\QFu\}$, and $\Pr(E(S_3)\cap\CQ_{Bok})=\frac{1}{9}$, $\Pr(E(S_3)\cap\CQ_{Dok})=0$,
    $\Pr(E(S_3)\cap\CQ_{Fok})=\frac{1}{3}$.
    
 We use (\ref{CD2}) to calculate the values of $\CD$:
 
  \NI$\CD(S_1)=-2\frac{2}{9}\log\frac{2}{9}-\frac{1}{3}\log\frac{1}{3}+\frac{7}{9}\log\frac{7}{9}\approx1.21$
  
\NI$\CD(S_2)=-\frac{2}{9}\log\frac{2}{9}-\frac{1}{3}\log3+\frac{5}{9}\log\frac{5}{9}\approx0.54$ 
  
\NI$\CD(S_3)=-\frac{1}{9}\log\frac{1}{9}-\frac{1}{3}\log3+\frac{4}{9}\log\frac{4}{9}\approx0.36$ 

For bigger subsets $S$ of $\QBt$ we get $\CD(S)=0$. 

The sequence of $\CD(S_k)$, above corresponds to the values of function $\deI(\QBt,k)$ from section~\ref{EntWghtAbstr} for $k=1,2,3$.

Notice that $\CD(S_i)=\max\{\CD(S): S\; \textrm{is a subset of}\; \QB3\wedge|S|=i\}$, the sequence 
$\CD(S_1),\CD(S_2),\CD(S_3)$ shows speed of convergence of entropic weight to $0$ for $\QB3$.

\
We formulate these observations in an abstract form.

\

  \section{Abstract Definition of Informativeness}\label{EntWghtAbstr}
  
  Notations:
    
  $\bullet$ $\CF$ a set (it is a set of formulas that is not made explicit above); for generality we may
  treat formulas modulo some equivalence relation but we do not use this option, just mention it.
  
  $\bullet$  $\CG\subset\CF$  is a subset of $\CF$ (these are goals, it corresponds to 
  $\{\PhI(\alI)\}_{\alI}$ above),  
  
  \Hc{0.4}$M=|\CG|$. 
  
  $\bullet$  $\CQ$ a set of subsets of $\CF$ (the set $\CQ$ of proofs above); 
  each $Q\in\CQ$ contains exactly one element of $\CG$,  its goal, and (for simplicity) each element of $\CG$ belongs to some $Q\in\CQ$. 
  
  $\bullet$ $\CQ_{\phI}=\{Q\in\CQ:\phI\in{}Q\}$ for  $\phI\in\CG$ (sets $\CQ_{\alI}$ above). 
    
  $\bullet$ Probabilistic measure $\Pr$ on $\CQ$: $\displaystyle\Pr(\CQ_{\phI})=\frac{1}{M}$
  for $\phI\in\CG$ ;  
  
  \Hc{0.4}$\displaystyle\Pr(Q)=\frac{1}{M\cdot|\CG_{\phI}|}$ for $Q\in\CQ_{\phI}$.
  
  $\bullet$ Entropic weight.   
  For a $Q\in\CQ$ and $S\subseteq{}Q$ set $E(S)=\{Q'\in\CQ:S\subseteq{}Q'\}$ and
  \begin{equation}\label{CDa1}
 \CD(S)=
  -\sum_{\phI\in\CG}\Pr(E(S)\cap\CQ_{\phI})\log\frac{\Pr(E(S)\cap\CQ_{\phI})}{\Pr(E(S))},
  \end{equation}
  or equivalently
      \begin{equation}\label{CDa2}
\CD(S)=-\sum_{\phI\in\CG}\Pr(E(S)\cap\CQ_{\phI})\log\Pr(E(S)\cap\CQ_{\phI})+\Pr(E(S))\log\Pr(E(S)).
    \end{equation}
  
  $\bullet$ $\deI(Q,k)=\max\{\CD(S):S\subseteq{}Q\wedge|S|=k\}$ for $Q\in\CQ$. This function is non-increasing
  when $k$ grows, it may be used to characterize speed of convergence to certainty.
  
  $\bullet$ $\zeI(Q)=\min\{k:\deI(Q,k)=0\}$. Clearly, $0<\zeI(Q)\leq|Q|$. This is the minimal size of subsets 
  of a proof that guarantees certainty in the worst case.
  
  $\bullet$ $\frac{1}{|Q|}\sum_{i=1}^{i=|Q|}\deI(Q,i)$ is an average entropic weight of $Q$.
  
  $\bullet$ $\frac{1}{\zeI(Q)-1}\sum_{i=1}^{i=\zeI(Q)-1}(\deI(Q,k)(i)-\deI(Q,k)(i+1)$ is an average speed of convergence to certainty. 
  
  \begin{prop}\label{CDprop}\;
    
   (D1) $\CD(\ES)=\log{}M$ (maximal uncertainty)
  
   (D2) $\CD(S)=0$ for any $\phI\in\CG$ and any $S\subseteq\CQ_{\phI}$ such that $E(S)\subseteq\CQ_{\phI}$
    
  \hspace{1cm}(maximal certainty)
  
   (D3) $\CD(S)$ (in fact, $\CD(E(S))$) is non-increasing when its argument $S$ grows: 
   
      \Hc{1}if $S\subseteq{}S'$ then $\CD(S)\geq\CD(S')$ 
      
      \hspace{1cm}(the uncertainty does not grow up with getting  
      more information).  
  \end{prop}
  
  \NBF{Proof}
  
   (D1). Indeed,  $E(\ES)=\CQ$, $\Pr(\CQ)=1$, 
   
   \NI$\CD(\CQ)=-\sum_{\alI}\Pr(\CQ_{\alI})\log\frac{\Pr(\CQ_{\alI})}{1}=
     -\sum_{\alI}\frac{1}{M}\log\frac{1}{M}=\log{}M$.
     
  (D2). Take $E(S)\subseteq\CQ_{\alI}$. Then $\CD(S)=-\Pr(E(S))\log\frac{E(S)}{E(S)}=0$, 
  
    \NI$E(S)\cap\CQ_{\beI}=\ES$ for $\beI\neq\alI$ and thus $\Pr(E(S)\cap\CQ_{\beI})=0$.
  
  Proof of (D3) can be done along the lines of the proof of similar property for the entropic weight introduced in \cite{Sli:2020:ECA}.
  
  Take any function of continuous time $S(t)\subseteq\CF$ such that $S(t_0)\subseteq{}S(t_1)$ for $t_0\leq{}t_1$.
  Then $E(S_1)\subseteq{}E(S_0)$ 
  
  Let $x_{\phI}(t)$ be a differentiable function, non-increasing when $t$ goes from $t_0$ to $t_1$,  such that
  \NI$x_{\phI}(t_0)=\Pr(E(S(t_0))\cap{}\CQ_{\phI})$ and $x_{\phI}(t_1)=\Pr(E(S(t_1)\cap{}\CQ_{\phI})$. Clearly, such a function exists and even can be easily constructed. 
  
  \NI{}We have  
  $\sum_{\phI}x_{\phI}(t_j)=\Pr(E(S(t_j)))$, $j=0,1$.
  
  Set   
  $p(t)=-\sum_{\phI}x_{\phI}(t)\log{}x_{\phI}(t)+
  \big(\sum_{\phI}x_{\phI}(t)\big)\log\big(\sum_{\phI}x_{\phI}(t)\big)$. 
  
  Then from (\ref{CDa2}) we see that $p(t_j)=\CD(S(t_j))$.
  
  We have $0\leq{}x_{\phI}(t)\leq\frac{1}{M}$ and $0\leq\sum_{\phI}x_{\phI}(t)\leq1$. 
  
  Assume that $S(t_0)$ is not empty, otherwise (D3) is trivial because of (D1). In this case 
  $0<x_{\phI}(t_1)\leq{}x_{\phI}(t)\leq{}x_{\phI}(t_0)$. Take derivative of $p(t)$ over $t$ (recall that
  $\log{}z=\frac{\ln{}z}{\ln2}$):
        \begin{eqnarray}
p'(t)=-\sum_{\phI}\,\Big(x_{\phI}'\log{x_{\phI}}+x_{\phI}\frac{x_{\phI}'}{x_{\phI}\cdot\ln2}\Big)+\nonumber\\
 \Big(\sum_{\phI}\,x_{\phI}'\Big)\log\Big(\sum_{\phI}\,x_{\phI}\Big)+
 \Big(\sum_{\phI}\,x_{\phI}\Big)\frac{\Big(\sum_{\phI}\,x_{\phI}'\Big)}{\Big(\sum_{\phI}\,x_{\phI}\Big)\ln2}=\nonumber\\
 -\sum_{\phI}\,\Big(x_{\phI}'\log{x_{\phI}}+\frac{x_{\phI}'}{\ln2}\Big)+
 \sum_{\phI}\,\Big(x_{\phI}'\log\Big(\sum_{\phI}\,x_{\phI}\Big) +
 \frac{x_{\phI}'}{\ln2}\Big)
 =\sum_{\phI}\,x_{\phI}'\cdot\log\frac{\Big(\sum_{\phI}\,x_{\phI}\Big)}{x_{\phI}}\label{eqnDerivCD}
      \end{eqnarray}
The functions $x_{\phI}$ are non-increasing, thus $x_{\phI}'\leq0$. As $\sum_{\phI}\,x_{\phI}\geq{}x_{\phI}$
 the value of (\ref{eqnDerivCD}) is non-positive, hence $p(t)$ is non-increasing when $S(t)$ increases. QED. 
  
  \
  
  \NBF{Remark.} With respect to sets $E(S)$ the function $\CD(E(S))$ is non-decreasing: if 
  $E(S)\subseteq{}E(S')$ then $\CD(E(S))\leq\CD(E(S'))$. When $S$ grows from $\ES$ to $\CQ_{\phI}$ the values 
  of $\CD(S)$ decrease from $\log{}M$ to $0$, so this decreasing is 'strict on the whole'.

\bibliographystyle{plain}
\bibliography{RelateInformProof}
 \end{document}